\documentclass[letterpaper]{article} 
\usepackage{aaai25}  
\usepackage{times}  
\usepackage{helvet}  
\usepackage{courier}  
\usepackage[hyphens]{url}  
\usepackage{graphicx} 
\usepackage{natbib}  
\usepackage{caption} 
\usepackage{algorithm}
\usepackage{algorithmic}
\usepackage{amsmath}
\usepackage{booktabs}
\usepackage{multirow} 
\usepackage{tabularx}
\usepackage{xcolor}
\usepackage{colortbl}
\definecolor{myred}{HTML}{FF4500}
\definecolor{myyellow}{HTML}{F4CD00}
\definecolor{mygreen}{HTML}{62B69C}
\definecolor{myblue}{HTML}{0066A0}
\definecolor{mygray}{HTML}{EEEEEE}

\usepackage{newfloat}
\usepackage{listings}
\DeclareCaptionStyle{ruled}{labelfont=normalfont,labelsep=colon,strut=off} 
\floatstyle{ruled}
\newfloat{listing}{tb}{lst}{}
\floatname{listing}{Listing}
\title{Multi-View Empowered Structural Graph Wordification\\ for Language Models}

\author{
    Zipeng Liu\textsuperscript{\rm 1,2}\thanks{Equal contribution.},
    Likang Wu\textsuperscript{\rm 1,2}\footnotemark[1],
    Ming He\textsuperscript{\rm 3},
    Zhong Guan\textsuperscript{\rm 1,2},
    Hongke Zhao\textsuperscript{\rm 1,2}\thanks{Corresponding author.},
    Nan Feng\textsuperscript{\rm 1,2}
}

\affiliations{
    \textsuperscript{\rm 1}College of Management and Economics, Tianjin University,
    \textsuperscript{\rm 2}Laboratory of Computation and Analytics of Complex Management Systems (CACMS), Tianjin University
    \textsuperscript{\rm 3}AI Lab, Lenovo Research
}

\begin{document}

\renewcommand{\thefootnote}{\fnsymbol{footnote}}

\maketitle

\begin{abstract}
Significant efforts have been dedicated to integrating the powerful Large Language Models (LLMs) with diverse modalities, particularly focusing on the fusion of language, vision and audio data. However, the graph-structured data, which is inherently rich in structural and domain-specific knowledge, has not yet been gracefully adapted to LLMs. Existing methods either describe the graph with raw text, suffering the loss of graph structural information, or feed Graph Neural Network (GNN) embeddings into LLMs at the cost of losing explainable prompt semantics. To bridge this gap, we introduce an end-to-end modality-aligning framework for LLM-graph alignment: \textbf{D}ual-\textbf{R}esidual Vector Quantized-Variational Auto\textbf{E}ncoder, namely \textbf{Dr.E}. Our approach is purposefully designed to facilitate token-level alignment with LLMs, enabling an effective translation of the intrinsic `language' of graphs into comprehensible natural language. We also manage to enhance LLMs' more robust structural understanding of graphs by incorporating multiple views of the central nodes based on their surrounding nodes at various distances. Our experimental evaluations on standard graph tasks demonstrate competitive performance against other state-of-the-art (SOTA) approaches. Additionally, our framework ensures certain visual interpretability, efficiency, and robustness, marking the promising successful endeavor to achieve token-level alignment between LLMs and GNNs. Our code is available at: https://github.com/Timothy914/Dr.E.
\end{abstract}

\section{Introduction}\label{sec:intro}

Until recently, Large Language Models (LLMs)~\cite{brown2020language, touvron2023llama, achiam2023gpt} were primarily regarded as text-centric, limited to communication through dialogue and exhibiting inferior performance with visual inputs. However, since the emergence of ChatGPT-4o in mid-2024, multimodal, end-to-end pre-trained LLMs have taken center stage. These advanced models possess the capability to process visual and auditory inputs, enabling them to engage with humans in a more natural and fluid manner. This advancement is supported by significant research efforts~\cite{radford2021learning, yu2022scaling, tsimpoukelli2021multimodal}, which have successfully integrated linguistic, visual, and audio modalities. These studies demonstrate that a diverse and high-quality dataset greatly enhances the performance of LLMs. Despite these achievements, the integration of graph-structured data into LLMs has not seen comparable breakthroughs.

\begin{figure}[t]
\centering
\includegraphics[width=1\linewidth]{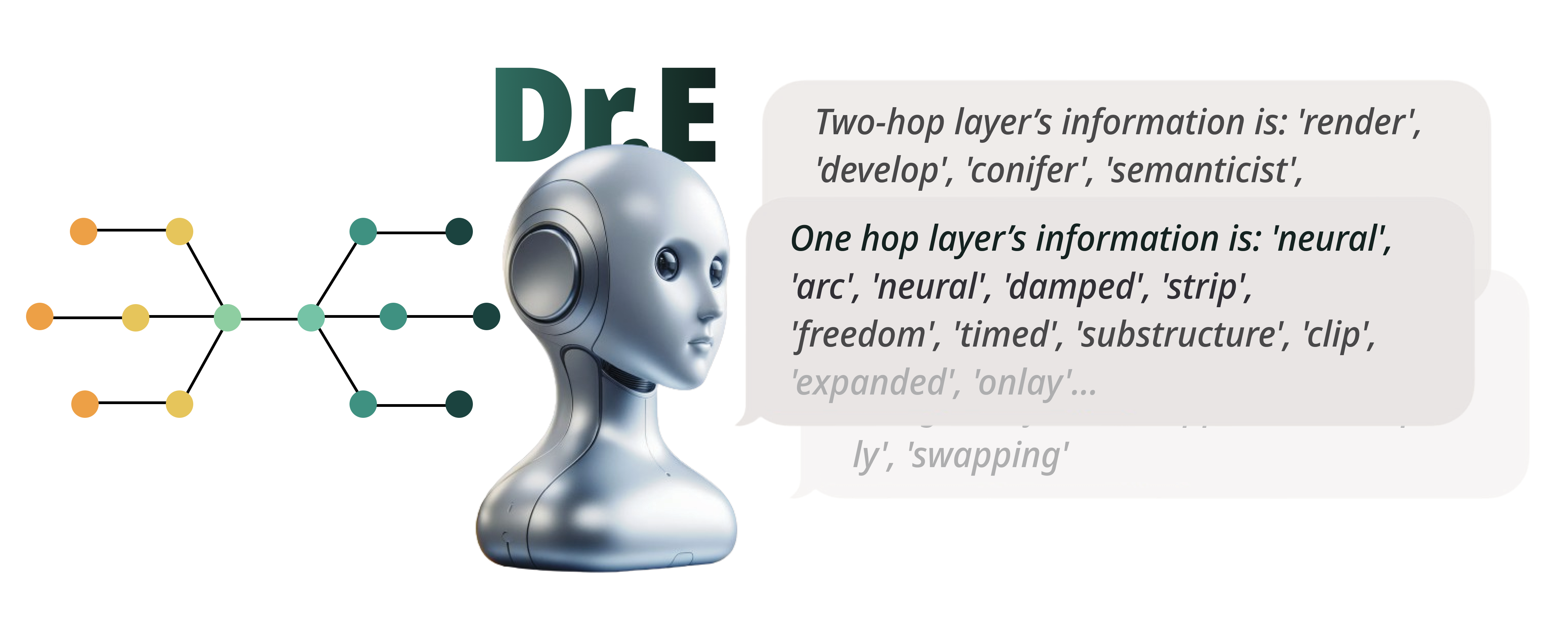}
\caption{A demonstration of our proposed framework, \textbf{Dr.E}, which serves effectively as a seamless interpreter, translating graphs (depicted on the left) into comprehensible natural language (as displayed on the right).}
\label{fig:overview}
\end{figure}

Graph data~\cite{kipf2016semi}, characterized by its domain-specific nature and diverse forms of features, is ubiquitous across various domains~\cite{zhou2020graph, wu2020comprehensive, wu2023survey, wu2023learning, zhao2023cross1, zhao2023sequential, zhao2024comprehensive}. For instance, in citation networks~\cite{mccallum2000automating, wu2024supporting}, it is used to predict the category of a given paper or the citation relationship between two papers. In recommendation systems~\cite{zhao2023cross, zhao2024lane, zhao2024collaborative}, graph nodes represent items and users, and the system recommends items to users by mining the underlying relationships among them. Graph data usually lacks textual attributes, which poses challenges to constructing a uniform input representation for LLMs. Traditional graph-based methods~\cite{spitzer2001principles, wu2020comprehensive}, have struggled to create a unified embedding space that aligns well with LLMs. Current methods~\cite{ye2023natural, wu2024exploring} that integrate LLMs with graph data mostly employ one of two lines: it either directly leverages textual description of node information and graph structure, or it feeds the embeddings derived from GNNs into LLMs.

Language serves as a natural medium to describe graphs, since LLMs are pre-trained on massive textual datasets. Non-linguistic features often elude interpretation through verbal descriptions and, as a result, cannot be processed effectively by LLMs. An alternative strategy involves directly integrating graph embeddings into LLMs, but this approach does not fully exploit the \textbf{linguistic interpretability} of LLMs. Furthermore, within the graph convolution process, there is a risk of losing \textbf{layer-specific information}, which can be detrimental to the ability of LLMs to grasp layer-distinguishable structural knowledge effectively. \underline{Contrary to existing methods}, we propose a novel solution that leverages the inherent linguistic processing capabilities of LLMs. By aligning graph embeddings with the LLMs’ vocabulary at different stages during the aggregation process, our approach enables LLMs to interpret graph structures linguistically. Given the intricacy of practical semantic information and the variability of graph structures, we have developed a codebook-driven method that employs multiple structural tokens (codes) to represent a single \textbf{view} of a graph's structure. This approach effectively captures structural nuances through the iterative generation of new codes. In addition, considering the ever-changing graph structures, in order to obtain more stable structural tokens, we carefully deploy a structural encoder that focuses on multiple hop-different views, thereby achieving more robust graph-language model alignment and reducing training bias.

Therefore, in this paper, we introduce the \textbf{D}ual-\textbf{R}esidual Vector Quantized-Variational Auto\textbf{E}ncoder (\textbf{Dr.E}), our `graph language' translator designed to seamlessly map structural graph data to tokens compatible with LLMs and subsequently leverage LLM as the predictor for evaluation tasks. The main contributions can be summarized as follows:
\begin{itemize}
  \item Our proposed model presents a promising effort to achieve Token-Level Alignment between GNNs and LLMs, setting a new standard for the integration of graph data in linguistic learning environments.
  \item By implementing intra-layer and inter-layer residuals, we ensure the preservation of layer-specific perspectives through Multi-View Structural Enhancement, allowing for more robust information transmission from the surrounding nodes to the center.
  \item We evaluate \textbf{Dr.E} on several real-world datasets, where it notably surpasses powerful GNN-based and LLM-based methods under various scenarios, demonstrating the effectiveness of our approach.
\end{itemize}

\section{Background}

\subsection{Vector Quantized Variational AutoEncoder}

Among research on modality fusion within the field of computer vision, the Vector Quantized-Variational AutoEncoder (VQ-VAE)~\cite{van2017neural} has made an appropriate choice of architecture. This model can be compared to a communication system, comprising an encoder that transforms observed data \(\mathbf{x}\) into a sequence of discrete latent variables \(\mathbf{e}\), which constitute the central codebook. A decoder then reconstructs the original observed data \(\mathbf{x}\) from these discrete variables \(\mathbf{e}\), formulated as:
\begin{equation}
\label{eq:1}
\text{Quantize}(E(\mathbf{x})) = 
\mathbf{e}_i,\quad i = \arg\min\limits_j \|\,E(\mathbf{x})-\mathbf{e}_j\|,
\end{equation}
where $E(\mathbf{x})$ denotes the embeddings after non-linear mapping of the input $\mathbf{x}$ by the encoder, and the quantization process aims to find the nearest vector in the codebook. The decoder then reconstructs the picture (or graph) through another non-linear mapping function. The overall objective of the standard VQ-VAE is described as:
\begin{equation}
\begin{split}
\label{eq:2}
\mathcal{L}(\mathbf{x}, D(\mathbf{e})) &= \|\mathbf{x} - D(\mathbf{e})\|_2^2 \\
&\quad + \|s g[E(\mathbf{x})] - \mathbf{e}\|_2^2 \\
&\quad + \beta\|s g[\mathbf{e}] - E(\mathbf{x})\|_2^2,
\end{split}
\end{equation}
in which $\|\mathbf{x}-D(\mathbf{e})\|_2^2$ is used to update the parameters of the decoder, $\|s g[E(\mathbf{x})]-\mathbf{e}\|_2^2$ represents the codebook loss, ensuring that the codes are close to the encoder's output, and $\|s g[\mathbf{e}]-E(\mathbf{x})\|_2^2$ is the commitment loss, encouraging the encoder to generate outputs that are close to the codes. Here, \( s g[\cdot] \) denotes the stop-gradient operation, and \(\beta\) is a hyperparameter that balances the commitment loss. The discrete codebook, on the one hand, helps the model learn distinct aspects of the data and explicitly represents these attributes in the latent space, leading to better disentanglement of features. On the other hand, it closely resembles the structure of LLMs' token embeddings, both of which do not require complex continuous optimization. Leveraging the latent expressive capabilities of the codebook in VQ-VAE, efforts have been directed toward aligning visual tokens with LLMs' vocabularies. For example, LQAE~\cite{liu2024language} employed a RoBERTa-augmented VQ-VAE model in few-shot learning settings. Following this, the CLIP~\cite{radford2021learning} model has been utilized to enhance visual and textual alignment in SPAE~\cite{yu2024spae} and V2L-tokenizer~\cite{zhu2024beyond}.

Despite these advancements in aligning visual and textual modalities, no pre-trained language-graph contrastive model similar to CLIP has emerged, given the unique nature of graph-structured data. Within the domain of GNNs, VQ-VAE has been adapted to facilitate knowledge distillation from GNNs to Multilayer Perceptrons (MLPs)~\cite{yang2023vqgraph}. TIGER~\cite{rajput2024recommender} employs a residual approach inspired by RQ-VAE~\cite{lee2022autoregressive} to conduct sequential generative retrieval. Currently, there is no recent work that combines the VQ-VAE architecture with the integration of LLM and GNN domains. Given that LLMs produce new tokens sequentially, we adopted a residual generation mechanism akin to that used in RQ-VAE to emulate the process of token generation. Moreover, because graph-based VQ-VAE architectures struggle to effectively capture the structural information of nodes within graphs, we introduced a hierarchical code generation approach—an innovative method specifically tailored for the graph domain.

\subsection{Intergration of LLM and GNN}

\begin{figure*}[t]
\centering
\includegraphics[width=0.98\textwidth]{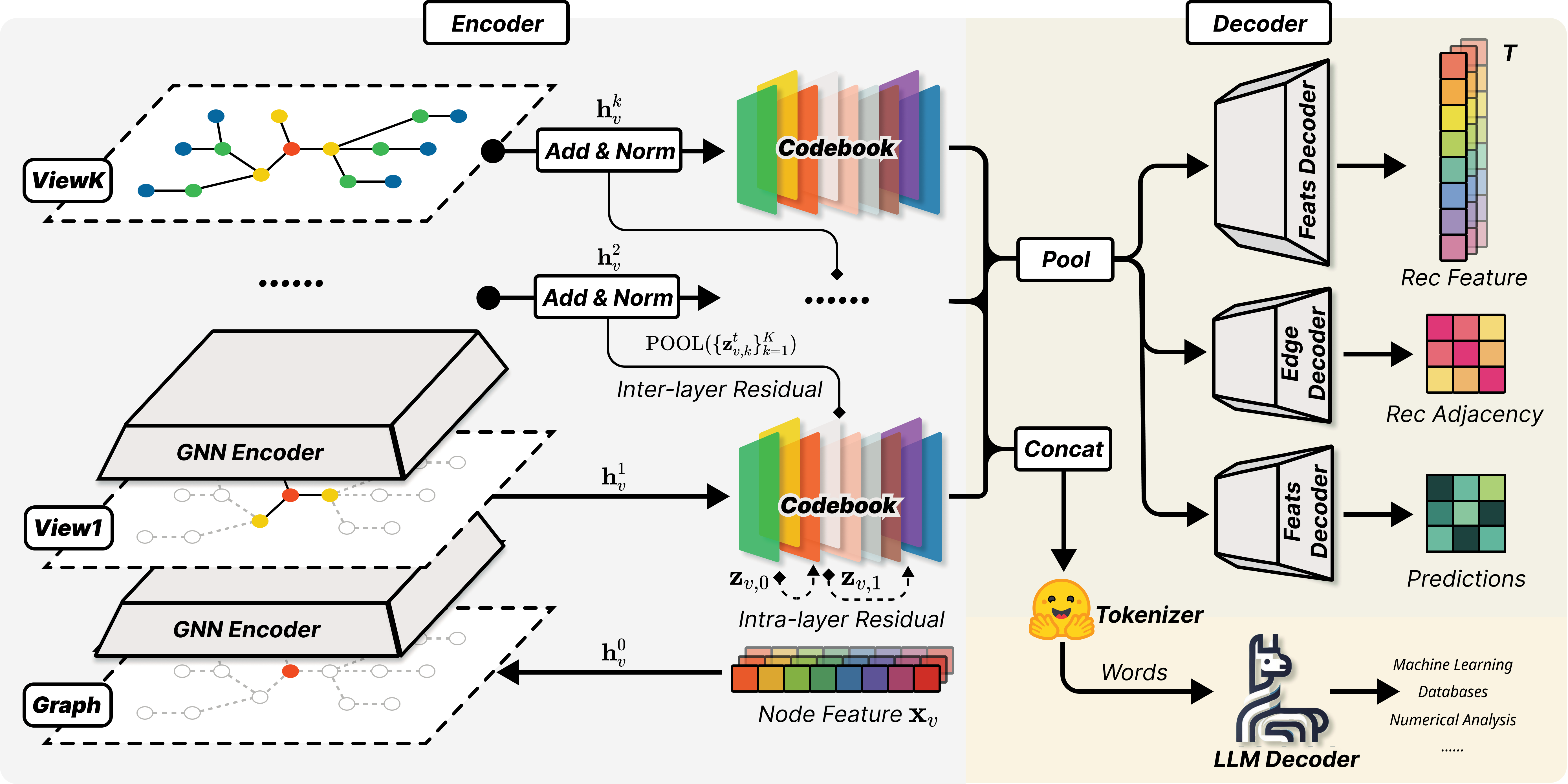} 
\caption{The overall framework of \textbf{Dr.E} encompasses a modified RQ-VAE architecture, where the encoder is a GNN module that directly processes the raw features of nodes in the graph, and the decoder is an LLM decoding codes' embeddings back to labels. We also incorporate additional features, labels, and adjacency matrix reconstruction to facilitate the training process. The token embeddings of the LLM serve as a critical codebook, bridging the encoder and the decoder seamlessly.}
\label{fig:framework} 
\end{figure*}

It can be broadly classified into two main types of approaches for integrating LLMs and GNNs~\cite{chen2024exploring}. The first category involves utilizing LLMs as enhancers to augment textual information, which is subsequently leveraged by GNNs for downstream tasks such as node classification. A representative example in this category is TAPE~\cite{he2023harnessing}, which employs LLMs to generate pseudo-labels and explanatory content. These explanations elucidate the logical connections between textual features and their corresponding labels. Based on these explanations, textual features and enhanced textual features are derived from both the original and enriched text attributes, respectively, serving as the initial node features for the GNNs. SimTeG~\cite{duan2023simteg} on the other hand replaces the original embeddings from Pre-Trained Language Models (PLMs) with those obtained through Parameter-Efficient Fine-Tuning (PEFT), while Graphformers~\cite{yang2021graphformers} promotes co-training between PLMs and GNNs through mutual generation of embeddings.

Instead of merely utilizing LLMs as annotators, it is more advantageous to engage them as predictors, where the final output of the model is presented entirely in natural language. An early attempt was to inject the graph embeddings directly into the prompt, thereby enabling LLMs to process node features, a technique employed by InstructGLM~\cite{ye2023natural} through instruction-tuning of the LLMs. Later, GraphGPT~\cite{tang2024graphgpt} sought to add new tokens into LLMs vocabulary through a dual-stage tuning process. Methods such as LLaGA~\cite{chen2024llaga}, which avoided fine-tuning the LLMs, instead employ a projector to map graph embeddings to token embeddings, subtly encoding graph structural information. In contrast, the UniGraph model~\cite{he2024unigraph} not only trains a graph encoder but also fine-tunes an LLM to execute general GNN tasks using large language models. 

However, these methods do not account for the linguistic characteristics that the graph itself might possess when feeding embeddings into the LLM. Consequently, we chose a more refined approach: directly translating the embeddings obtained through GNN aggregation into natural language, thereby preserving the linguistic essence of the LLM to the greatest extent possible.

\section{Methodology}

With the goal of infusing LLMs with the structural knowledge embedded in graphs, our proposed architecture adopts an end-to-end fine-tuning approach for better alignment between GNNs and LLMs. To bridge the continuous GNN embeddings with the discrete vocabulary of LLMs, we have carefully constructed a model architecture based on RQ-VAE, utilizing GNNs as the encoder and LLMs as the decoder. The input is simply composed of nodes' features in a graph, and predictions are directly yielded for downstream tasks in natural language.

\subsection{Multi-View Structural Enhancement}

 The most distinctive design within the GNN module involves preserving the multi-view of a graph, which proves to be effective in encoding the structural information of the central node and consequently improves the performance of the LLMs. During the training process, we sample a subgraph $\mathcal{G}=\left(\mathcal{V}, \mathcal{E}\right)$ with node features $\left\{\mathbf{x}_v, \forall v \in \mathcal{V}\right\}$,  following the procedure outlined in GraphSAGE~\cite{hamilton2017inductive}. Given a node $v$ with its embedding $\mathbf{h}_v$, our goal is to equip the LLMs with an understanding of the specific positional information of a node within a graph. This involves acquiring layer-specific information about the relevant neighboring nodes of a central node at each aggregation step. To achieve this, we introduce a smaller subgraph \(\mathcal{G}_d = (\mathcal{V}_d, \mathcal{E}_d)\) consisting of nodes that are precisely within \(d\) hop(s) away from a selected central node. Here \(\mathcal{V}_d \subseteq \mathcal{V}\) denotes the set of these nodes and \(\mathcal{E}_d \subseteq \mathcal{E}\) includes all edges connecting them. For example, $\mathcal{G}_2$ includes the nodes directly connected to the central node $v$, denoted as $\mathcal{N}(v)$ and the nodes directly connected to $\mathcal{N}(v)$, denoted as $\mathcal{N}(\mathcal{N}(v))$. This subgraph \(\mathcal{G}_d\) represents a level of \textbf{view} in our framework. A total of $D$ such subgraph views are sequentially fed into the downstream pipeline, illustrated as:
\begin{equation}
\label{eq:3}
Predictions=LLM(\mathcal{G}_1, \mathcal{G}_2, \dots, \mathcal{G}_D).
\end{equation}

For graph-structured data, the information about node connections is embedded within each aggregation process. By isolating each view, we effectively enhance the structural information of the graph, enabling the embeddings obtained through aggregation to concentrate more on the information pertinent to that particular subgraph level. This allows the LLMs to gain deeper insights into the connectivity between nodes at that level of the view. Through an ablation study, we have demonstrated that the information from each level of view contributes to improving the LLMs' understanding of the graph structure, a technique we term \textbf{Multi-View Structural Enhancement}. The empirical results show significant improvements in model performance, validating the effectiveness of our technique.

\subsection{Token-Level Alignment}

Within a standard VQ-VAE architecture, a discrete latent variable $\mathbf{e}_i$ is selected to represent an observation, denoted as $\mathbf{x}$. This selection process is facilitated by an encoder-decoder pair, as formalized in Equations~\ref{eq:1} and~\ref{eq:2}. In our scenario, to select tokens that LLMs can interpret, an intuitive approach involves retrieving the most similar embedding $\mathbf{e}_c$ from a codebook $\mathcal{C}$, ensuring effective representation of the node while also conveying semantic meaning. The most direct solution is substituting the dynamic codebook from the original VQ-VAE with the token embeddings from LLMs. This method has been validated by LQAE~\cite{liu2024language}, formulated as:
\begin{equation}
\label{eq:4}
\mathcal{C}=\{(c, \mathbf{e}_c \mid c \in \mathcal{O}\},
\end{equation}
where $\mathbf{e}_c$ represents the token embedding for index $c$, and $\mathcal{O}$ is the subset of the LLM's vocabulary. Through this substitution, we achieve the transition from graph embeddings to language embeddings. During the fine-tuning phase of the LLM, natural language is used as input. As illustrated in Table~\ref{tab:prompt}, we provide potential node classification outcomes and the multi-view information of the central node, derived from the previous section, as prompts to the LLMs. Consequently, the model updates the parameters of both the GNN and the LLM synchronously, guided by the LLM's loss function.

\subsection{Dual-Level Residue}

To enhance the stability and performance of \textbf{Dr.E}, we introduce additional Intra-Layer and Inter-Layer Residual modules. The Intra-Layer Residual module boosts the expressive ability of the codes selected from the codebook, whereas the Inter-Layer Residual module mitigates over-smoothing issues during GNN aggregations. The specific designs of these modules are detailed as follows.

\subsubsection{Intra-Layer Residue}

Using a single token per view proves insufficient for capturing the information embedded within neighboring nodes, thereby reducing performance. To address this limitation, we employ a residual quantization technique to generate a sequence of codes. Initially, we select the closest code to the output of the encoder from the codebook $\mathcal{C}$, and then subtract the code embedding from the original output to form a new output. This process is repeated iteratively to generate new codes, formalized as:
\begin{equation}
\label{eq:5}
\mathbf{z}_{v,k} = \arg \min_{\mathbf{e}_c} \left\| \mathbf{h}_{v,k-1} - \mathbf{z}_{v,k-1} - \mathbf{e}_c \right\|,
\end{equation}
in which $\mathbf{z}_{v,k}$ denotes the embedding after quantization. Subsequently, the update equation for $\mathbf{h}_{v,k}$ is given by:

\begin{equation}
\label{eq:6}
\mathbf{h}_{v,k} = \mathbf{h}_{v,k-1} - \mathbf{z}_{v,k},
\text{ for   } k = 1, \ldots, K,
\end{equation}
where $k$ represents the index of the code selected from the codebook, and $K$ is the total number of codes used to quantize a node. The initial conditions are $\mathbf{h}_{v,0} = \mathbf{h}_v$ and $\mathbf{z}_{v,0} = \mathbf{0}$. The $K$ codes combined together represent the node $v$, with the intra-layer residual $\mathbf{r}_{v,k}$ at step $k$ defined as $\mathbf{h}_{v,k} - \mathbf{h}_{v,k-1}$.

\subsubsection{Inter-Layer Residue}

After obtaining quantized embeddings via Equation~\ref{eq:5} and Equation~\ref{eq:6}, the oversmoothing problem, typical for GNNs, occurs. To mitigate this, We apply a pooling operation on these embeddings, which are then concatenated with the original node embeddings prior to aggregating the embeddings of surrounding nodes $\mathcal{N}(v)$. The forward propagation is defined as:
\begin{equation}
\begin{aligned}
\label{eq:7}
\mathbf{h}^{t+1}_v &= \sigma\Big(\mathbf{W}^t \cdot \textsc{CONCAT}\Big( \mathbf{h}_v^t \\
&\quad + \textsc{POOL}(\{\mathbf{z}^t_{v,k}\}_{k=1}^K), \mathbf{h}_{\mathcal{N}(v)}^t\Big)\Big),
\end{aligned}
\end{equation}
where $\forall u \in \mathcal{N}(v)$ denotes any neighbor $u$ of the central node $v$, and $t$ ranges from $1$ to $T$, with $T$ being the total number of convolution steps. Additionaly, $\mathbf{h}_v^0=\mathbf{x}_v$ and $\mathbf{z}_{v,k}^0=\mathbf{0}$. At this step, the inter-layer residue $\mathbf{r}^t_v$ is calculated as $\mathbf{r}^t_v=h^{t}_{v}-h^{t-1}_{v}$. More formally, the embeddings $\mathbf{h}^{t}_v$ at step $t$ can be seen as an abstraction of the embeddings $\mathbf{h}^{t+1}_v$ at step $t+1$. This is because $\mathbf{h}^{t+1}_v$ encapsulates information from nodes one additional hop away, while maintaining the same embedding size as $\mathbf{h}^{t}_v$. Furthermore, considering the utilization of multiple layers of views, the representation of node $v$, requires fewer codes due to the expanded representation space, which is the product of the number of codes at each step $t$, shown as:
\begin{equation}
\label{eq:8}
\|\mathcal{E}\| = \|\mathcal{C}\|^T,
\end{equation}
in which $\|\mathcal{E}\|$ denotes the representation space of \textbf{Dr.E}'s encoder and $\|\mathcal{C}\|$ represents the size of the codebook $\mathcal{C}$. Additionally, the dual-residual generation process mirrors the token generation mechanism during the inference phase of the LLMs. This enhances the representation of sequential information, as the newly generated code depends on the previously obtained codes, described as:
\begin{equation}
\begin{aligned}
\label{eq:9}
z_{v, k}^t &= \arg \max _{z_{v, k}^t} P\Big(z_{v, k}^t \mid z_{v, 1}^1, z_{v, 2}^1, z_{v, 3}^1 \ldots, \\
&\quad  z_{v, K}^1,\ldots, z_{v, k-1}^t\Big).
\end{aligned}
\end{equation}

\subsection{Auxiliary Reconstruction Loss}

 Mean Squared Error (MSE) Loss and Cross-Entropy (CE) Loss are employed for feature reconstruction and label prediction, respectively. Given the imbalance between positive and negative samples resulting from the edge sampling process during adjacency reconstruction, we modify the Binary Cross-Entropy loss to a Weighted Binary Cross-Entropy loss, as implemented by VGAE~\cite{kipf2016variational}. This loss function is defined as:
\begin{equation}
\label{eq:10}
\begin{split}
\mathcal{L}_{\text{adjacency}} = & -\frac{1}{N} \sum_{i=1}^N \Bigl[ w_i y_i \log \left(\hat{y}_i\right) \\
& + \left(1-y_i\right) \log \left(1-\hat{y}_i\right) \Bigr],
\end{split}
\end{equation}
where $N$ represents the total number of sampled edges within a batch. $y_i, \hat{y}_i \in [0,1]$ denotes the ground truth and predictions for the existence of an edge, respectively. The weight $w_i=\frac{N}{2*N_i}$ is assigned to the $i$-th sample to adjust its contribution to the loss function and mitigate the effects of sample imbalance, in which $N_i$ is the number of edges of the same class as the $i$-th sample.

\section{Experiments}

\subsection{Experimental Settings}

\subsubsection{Datasets}

To evaluate the efficacy of our framework, \textbf{Dr.E} is tested on three benchmark datasets: Cora~\cite{mccallum2000automating}, PubMed~\cite{sen2008collective}, and OGBN-Arxiv~\cite{hu2020open}. These datasets are chosen for their widespread use in evaluating methods combining LLM and GNN. Each dataset represents a citation network, with the primary tasks being node classification. Specifically, the tasks involve predicting the category label of a target node. Notably, while these datasets include information such as node titles, our model does not utilize any auxiliary textual data. This design decision ensures that our method is generalizable to any dataset lacking textual information. Statistical details for each of the three datasets are provided in Table~\ref{tab:dataset}. We adhere to the dataset splits commonly employed by other methods, such as those detailed in \cite{he2023harnessing}.

\begin{table}[h]
\caption{The statistical information of the datasets is shown below, where `\#Node' and `\#Edge' represent the number of nodes and edges, respectively, and `Sparsity‱' indicates the density of the graph, computed as $S = 1 - \frac{|E|}{n(n-1)}$.}
\centering
\begin{tabular}{l|ccc}
\toprule
\textbf{Dataset}  & \textbf{\#Node} & \textbf{\#Edge} & \textbf{Sparsity‱} \\
\midrule
Cora & 2,708 & 5,429 & 14.8120 \\
Pubmed & 19,717 & 44,338 & 2.2811 \\
Arxiv & 2,449,029 & 61,859,140 & 0.2067 \\
\bottomrule
\end{tabular}
\label{tab:dataset}
\end{table}

\subsubsection{Implementation Details}

To enhance the reproducibility of our experiments and ensure a fair comparison with prior work, we utilize the Llama2-7B as our LLM decoder, which is also widely adopted in various other studies. However, this choice introduces biases when compared with other studies that use variants of LLaMA or other LLMs as foundational models, such as InstructGLM~\cite{ye2023natural}, GraphGPT~\cite{tang2023graphgpt}, LLaGA~\cite{touvron2023llama}, and others. Our model incorporates three graph convolutional layers as an encoder to capture three views of representation when encoding a node into tokens. To prevent zero values from dominating the embeddings, we include an activation function and a quantizing module after dropout regularization. The convolution blocks are uniform and scalable, allowing them to be stacked multiple times. We employ cosine similarity to identify the nearest token, a method that has been shown to outperform a language model (LM) head layer in terms of convergence speed of the model. We implement LoRA PEFT adjustments for Llama2-7B and establish two distinct learning rates for the GNN encoder and LLM decoder, set at $1\times10^{-3}$ and $1\times10^{-4}$, respectively, with a weight decay of $5\times10^{-4}$. The hidden dimension for the SAGE convolution is 4096, matching the token embedding of Llama. Our experiments are conducted using 2 NVIDIA A800-SXM4-80GB GPUs.

\begin{table}[h]
\renewcommand{\arraystretch}{1.1}
\caption{We conduct comparative experiments for \textbf{Dr.E} on node classification task. The first horizontal section compares models purely based on GNNs; the second section covers Transformer-based graph models; and the third category includes methods that integrate graph models with LLMs. We compute the average performance of each model across three datasets and place the metrics in the last column. \textbf{Bold} signifies the best result across all methods, while an \underline{underline} highlights the second best result.}
\centering
\begin{tabular}{l|cccc}
\toprule
\textbf{Model} & \textbf{Cora} & \textbf{PubMed} & \textbf{Arxiv} & \textbf{Average} \\
\midrule
GCN & 0.8778 & 0.8031 & 0.7182 & 0.7997 \\
GraphSage & 0.8824 & 0.8881 & 0.7171 & 0.8292 \\
GAT & 0.8595 & 0.8328 & 0.7366 & 0.8096 \\
RevGAT & 0.8911 & 0.8850 & 0.7402 & 0.8338 \\
DRGAT & 0.8977 & 0.8962 & 0.7416 & 0.8452 \\
SGC & 0.8797 & 0.8735 & 0.7177 & 0.8236 \\
SAGN & 0.8919 & \underline{0.9517} & 0.7570 & 0.8669 \\
\cmidrule(lr){1-5}
Graphformer & 0.8044 & 0.7699 & 0.6725 & 0.7489 \\
Nodeformer & 0.8848 & 0.7958 & 0.6960 & 0.7922 \\
\cmidrule(lr){1-5}
InstructGLM & 0.8977 & 0.9105 & 0.7297 & 0.8459 \\
GLEM & 0.8856 & 0.9459 & 0.7580 & 0.8632 \\
GraphMAE2 & 0.8011 & 0.6983 & 0.7201 & 0.7398 \\
UniGraph & 0.8184 & 0.7433 & 0.7291 & 0.7636 \\
GraphEdit & \underline{0.9090} & 0.9409 & 0.7578 & 0.8692 \\
LLaGA & 0.8922 & 0.9503 & \textbf{0.7666} & \underline{0.8697} \\
\cmidrule(lr){1-5}
\textbf{Dr.E} & \textbf{0.9132} & \textbf{0.9670} &  \underline{0.7645} & \textbf{0.8815} \\
\bottomrule
\end{tabular}
\label{tab:main_result}
\end{table}

\subsubsection{Baselines}

In the comparative experiments, we select three categories of methods. The first category encompasses classical approaches that are purely based on GNNs, including GCN~\cite{kipf2016semi}, GraphSage~\cite{hamilton2017inductive}, GAT~\cite{velivckovic2017graph}, SGC~\cite{wu2019simplifying}, and SAGN~\cite{sun2021scalable}. The second category comprises Transformer-based graph models, such as Nodeformer~\cite{wu2022nodeformer} and Graphformer~\cite{wang2024graphformer}. The third category includes hybrid models that integrate LLMs with GNNs, including GLEM~\cite{zhao2022learning}, InstructGLM~\cite{ye2023natural}, GraphMAE2~\cite{hou2023graphmae2}, Unigraph~\cite{he2024unigraph}, GraphEdit~\cite{guo2024graphedit}, and LLaGA~\cite{chen2024llaga}. These experiments cover a sufficiently broad range to provide comprehensive insights into the comparative performance of these models.

\subsection{Main Evaluation}

Based on the results presented in Table~\ref{tab:main_result}, it is evident that \textbf{Dr.E} achieves state-of-the-art (SOTA) performance in terms of accuracy across the majority of the experimental settings. Interestingly, most LLM-based approaches demonstrate significant improvements over traditional graph-based methods. However, some of these methods leverage textual information associated with the nodes, which may have been utilized during the pre-training phase of LLMs on these datasets. In contrast, our experiments, which do not rely on textual node information, eliminate the potential for label leakage from pre-training and yield superior results, thereby substantiating the robustness and effectiveness of our approach. While our model performs less favorably than LLaGA on the OGBN-ArXiv dataset, it significantly outperforms other baseline methods on both the Cora and PubMed datasets. Compared with our model, LLaGA shows bias and insufficient generalization ability. Our robust performance across all three datasets reaches the obvious SOTA level.

\subsection{Model Analysis}

\subsubsection{Abalation Study}

To evaluate the contribution of each component of \textbf{Dr.E} to the overall model performance, we design a series of ablation experiments, detailed in Table~\ref{tab:abalation}. These experiments consist of multiple iterations. \textbf{Zero-Shot Prediction with Textual Information:} We begin by extracting the textual information (titles and extractions) of nodes and performing zero-shot predictions using Llama2-7B. This demonstrates the LLM's ability to comprehend textual attributes in graph data and classify papers based solely on their titles. \textbf{Sequential Training of GNN and Frozen LLM:} We then train the GNN encoder while keeping the LLMs' parameters frozen and updating only the GNN parameters. This sequential training improves the overall model performance, showing that a single GNN can align node embeddings with the input requirements of the language model to some degree.
\textbf{Fine-tuning LLM with LoRA:} Building on this, we incorporate Low-Rank Adaptation(LoRA) for fine-tuning the LLM, which leads to a significant improvement in performance. \textbf{Multi-View Enhancement:} Next, we apply the Multi-View Enhancement technique to structurally enhance the model's performance, resulting in a significant improvement in scores. \textbf{Quantization and Token Mapping:} To integrate semantic information into the LLM, we quantize the embeddings and map them to tokens within Llama2-7B. However, this results in a slight decrease in performance, suggesting the need for additional adjustments. \textbf{Dual-Layer Residual Mechanism and Codebook Refinement:} Finally, we introduce the dual-layer residual mechanism and preprocess the codebook by filtering out tokens that do not carry meaningful semantic information. With these improvements, the model achieves its best performance. These steps systematically evaluate the influence of each component on the overall effectiveness of \textbf{Dr.E}, highlighting critical aspects for optimizing the model's performance.

\begin{table}[t]
\centering\
\caption{Ablation studies on codebook with classification accuracy is evaluated on three datasets.}
\begin{tabular}{l|ccc}
\toprule
\textbf{Method} & \textbf{Cora} & \textbf{Pubmed} & \textbf{Arxiv}  \\
\midrule
Llama2-7B + Raw Text            & 0.6700  & 0.9075 & 0.5175  \\
Llama2-7B + GNN                 & 0.8319  & 0.8756 & 0.6513  \\
+ LoRA                          & 0.8824  & 0.9213 & 0.7235  \\
+ Multi-View                    & 0.9112  & 0.9524 & 0.7468  \\
+ VQ \& Frozen Codebook         & 0.8875  & 0.9482 & 0.7043  \\
+ Intra-layer Residual          & 0.9036  & 0.9598 & 0.7354  \\
+ Inter-layer Residual          & 0.9101  & 0.9655 & 0.7509  \\
+ Token Refinement              & 0.9132  & 0.9670 & 0.7645  \\
\bottomrule
\end{tabular}
\label{tab:abalation}
\end{table}

\subsubsection{Multi-View Analysis}

\begin{figure}[t]
\centering
\includegraphics[width=1\linewidth]{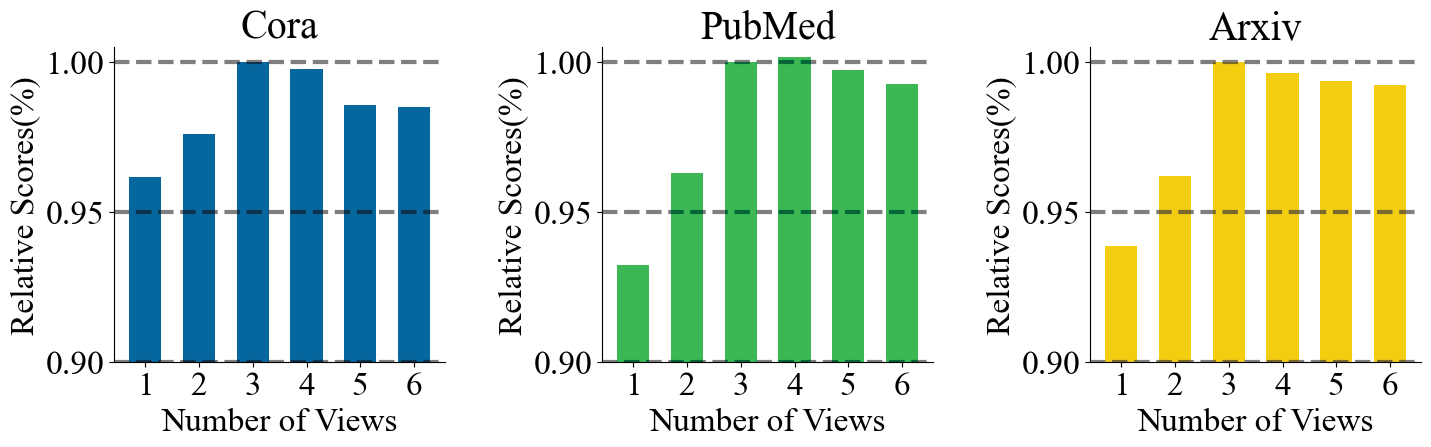}
\caption{We investigate the effect of the number of views on the model's performance. The x-axis represents the number of views used, while the y-axis shows the relative performance of the model under different numbers of views, normalized against the performance achieved with 3 views.}
\label{fig:multi-view}
\end{figure}

We analyze the impact of incorporating varying numbers of multi-view perspectives on the model's performance. We incrementally increase the number of views from one up to six additional perspectives. As illustrated in Figure~\ref{fig:multi-view}, the model’s performance consistently improves with the addition of more views, achieving its peak performance approximately at three views. This finding is consistent with general observations in GNN training. Based on these results, we opt for three distinct views as the optimal number for the central nodes, achieving a favorable balance between performance and efficiency.

\subsubsection{Codebook Analysis}

\begin{figure}[t]
\centering
\includegraphics[width=0.47\textwidth]{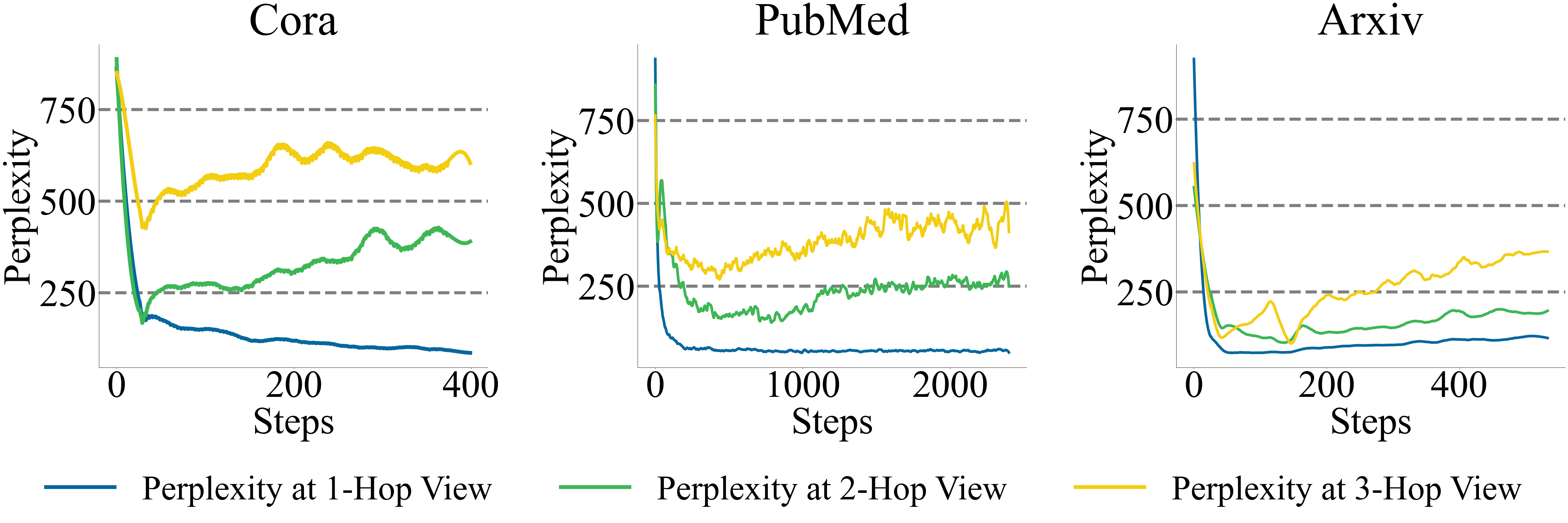} 
\caption{The figure above shows the perplexity for selection of tokens on each dataset. The blue line represents the perplexity of the 1-hop view of the codes, while the green and yellow lines represent the 2-hop and 3-hop views, respectively. Note that we apply a Savitzky-Golay filter when plotting the line graph to improve readability.}
\label{fig:perplexity} 
\end{figure}

To assess the effective utilization of the codebook during the encoding process, we calculate perplexity, defined as the exponential of the negative average log probability of the code distribution. This measure provides insight into how well the latent space is being utilized, which is a critical factor for model performance. The formula for perplexity is given by:
\begin{equation}
\label{eq:11}
\textsc{Perplexity} = \exp\left(-\sum_{i=1}^{\|\mathcal{C}\|} p_i \log p_i\right),
\end{equation}
where $|\mathcal{C}|$ represents the total number of codes in the codebook, and $p_i$ denotes the probability of the $i$-th code being used, typically estimated by the relative frequency of each code in the dataset. During the training process, the perplexity for each layer's view initially decreases to a minimum before rebounding to more stable values. Additionally, the perplexities across the three layers maintain a consistent ratio of 1:2:3, indicating that the amount of information aggregated increases as it propagates through successive layers as shown in Figure~\ref{fig:perplexity}. At the same time, the perplexity of the 1-hop view does not collapse to zero, indicating that our multi-view construction is sound and meaningful. For the central node, each layer's view encompasses distinct information pertinent to that specific layer.

\subsubsection{Case Study}
We also execute a case study, which involves analyzing the prediction for a real sample. In Table~\ref{tab:prompt}, we feed both the potential classification outcomes and the words represented by nodes into the LLM. The LLM then selects the most likely outcome based on the node information translated into natural language. This process reveals that while not every word carries positive label-related semantics, collectively, they contribute significantly to the final prediction. Although the words selected by our model do not necessarily convey practical semantic information, in some instances, \textbf{Dr.E} exhibits a certain degree of interpretability. As illustrated in Table~\ref{tab:prompt}, the terms `empirical', `case' and `statistically' are chosen to represent the graph structure for the central node, and the final classification for the node indeed reflects a `Case-Based' approach.

\begin{table}[t]
\centering
\caption{A real case for \textbf{Dr.E} illustrates how the embeddings of nodes within a graph are translated into the natural language that the LLMs can directly comprehend. The gray background section denotes the prompts input into the language model, the blue text signifies the actual classification outcomes, and the red text highlights words that carry information aggregated by the GNN.}
\label{tab:prompt}
\begin{tabularx}{\linewidth}{X}
\toprule
\textbf{Ground Truth} \\
\textbf{Title:} Case-Based Similarity Assessment: Estimating Adaptability from Experience \\
\textbf{Abstract:} Case-based problem-solving systems rely on similarity assessment to select stored cases whose solutions are easily adaptable to fit current problems... \\
\textbf{Label:} \textcolor{myblue}{\textbf{Case Based}} \\
\midrule
\cellcolor{mygray}\textbf{Prompt} \\
\cellcolor{mygray}\textbf{\textless User\textgreater:} Given a node, you need to classify it among `\textbf{Case Based}', `\textbf{Genetic Algorithms}'.... With the node's 1-hop information being `\textcolor{myred}{\textbf{justifiable}}', `\textcolor{myred}{\textbf{empirical}}', `\textcolor{myred}{\textbf{test}}'..., 2-hop information being `\textcolor{myred}{\textbf{assessment}}', `\textcolor{myred}{\textbf{case}}', `\textcolor{myred}{\textbf{empirical}}'..., 3-hop information being `\textcolor{myred}{\textbf{statically}}', `\textcolor{myred}{\textbf{combining}}', `\textcolor{myred}{\textbf{semantically}}', the node should be classified as: \\
\cellcolor{mygray}\textbf{\textless Assistant\textgreater:} \textcolor{myblue}{\textbf{Case Based}} \\
\bottomrule
\end{tabularx}
\end{table}



\section{Conclusion}

In this study, we present an end-to-end framework that seamlessly integrates graph-structured data with LLMs. Our framework leverages graph-structured information through Multi-View Structural Enhancement and {Token-Level Alignment, employing Intra-Layer and Inter-Layer Residue. This innovative approach effectively bridges the gap between GNNs and LLMs, enabling the translation of graph data into natural language at the token level while preserving both semantic and structural integrity. \textbf{Dr.E} demonstrates superior performance in GNN node classification tasks, showcasing robustness and efficiency in fine-tuning scenarios. However, handling extremely large and complex graphs presents challenges, and the current model is limited by its computational requirements. While our framework achieves notable results in comparative experiments, its computational demands remain a limitation. Consequently, the percentage improvement over other models, though evident, is not particularly significant. To better demonstrate our model's generalization capabilities on non-textual graphs, we plan to conduct experiments on additional datasets. Future research may also explore advanced tokenization techniques and scalable training methods to further enhance the performance of our framework.

\section{Acknowledgments}

This study was partially funded by the supports of National Natural Science Foundation of China (72471165, 72231004, 72101176).

\bibliography{aaai25}
\end{document}